\pdfoutput=1

\documentclass[11pt]{article}

\usepackage{EMNLP2022}

\usepackage{times}
\usepackage{latexsym}

\usepackage[T1]{fontenc}

\usepackage[utf8]{inputenc}

\usepackage{microtype}
\usepackage{hyperref}
\hypersetup{
    urlcolor=blue
    }

\usepackage{graphicx}
\usepackage{tabularx}
\usepackage{adjustbox}
\usepackage{booktabs}
\usepackage{multirow}
\usepackage{amsmath}
\usepackage{bbm}
\usepackage{amssymb}
\usepackage{cuted}
\usepackage{comment}
\usepackage{lipsum}
\usepackage{color}
\usepackage{colortbl}
\usepackage{xcolor}
\usepackage{relsize}
\usepackage{makecell}

\title{Rescue Implicit and Long-tail Cases: Nearest Neighbor Relation Extraction}

\author{
Zhen Wan \thanks{\quad This denotes equal contribution.}$\ \,^1$  \hspace{1em}
Qianying Liu \footnotemark[1]$\ \,^{1}$ \hspace{1em} \\
{\bf Zhuoyuan Mao$^1$ } \hspace{1em}
{\bf Fei Cheng$^1$ } \hspace{1em} 
{\bf Sadao Kurohashi$^1$ } \hspace{1em} 
{\bf Jiwei Li$^2$ }\\
$^1$ Kyoto University, Japan \hspace{1em}
\\
$^2$ Zhejiang University, China \hspace{1em}
\\
\texttt{\{zhenwan, ying, zhuoyuanmao\}@nlp.ist.i.kyoto-u.ac.jp} \\
\texttt{\{feicheng, kuro\}@i.kyoto-u.ac.jp} \\
\texttt{\{jiwei\_li\}@zju.edu.cn} \\
}
\begin{document}
\maketitle
\begin{abstract}

Relation extraction (RE) has achieved remarkable progress with the help of pre-trained language models. However, existing RE models are usually incapable of handling two situations: implicit expressions and long-tail relation types, caused by language complexity and data sparsity. In this paper, we introduce a simple enhancement of RE using $k$ nearest neighbors ($k$NN-RE). $k$NN-RE allows the model to consult training relations at test time through a nearest-neighbor search and provides a simple yet effective means to tackle the two issues above. Additionally, we observe that $k$NN-RE serves as an effective way to leverage distant supervision (DS) data for RE. Experimental results show that the proposed $k$NN-RE achieves state-of-the-art performances on a variety of supervised RE datasets, i.e., ACE05, SciERC, and Wiki80, along with outperforming the best model to date on the i2b2 and Wiki80 datasets in the setting of allowing using DS. \textcolor{black}{Our code and models are available at: \href{https://github.com/YukinoWan/kNN-RE}{https://github.com/YukinoWan/kNN-RE}.}
\end{abstract}

\section{Introduction}
Relation extraction (RE) aims to identify the relationship between entities mentioned in a sentence, and is beneficial to a variety of downstream tasks such as question answering and knowledge base population. Recent studies~\cite{zhang-etal-2020-minimize,Zeng_Zhang_Liu_2020,lin-etal-2020-joint,wang-lu-2020-two,cheng-etal-2020-dynamically,zhong-chen-2021-frustratingly} in supervised RE 
take advantage of 
 pre-trained language models (PLMs) and achieve SOTA performances by fine-tuning PLMs with a relation classifier. However, we observe that 
 existing
RE models are usually incapable of handling two RE-specific situations
: \textbf{implicit expressions} and \textbf{long-tail relation types}.

\begin{figure}[t]
    \centering
    \includegraphics[width=\linewidth]{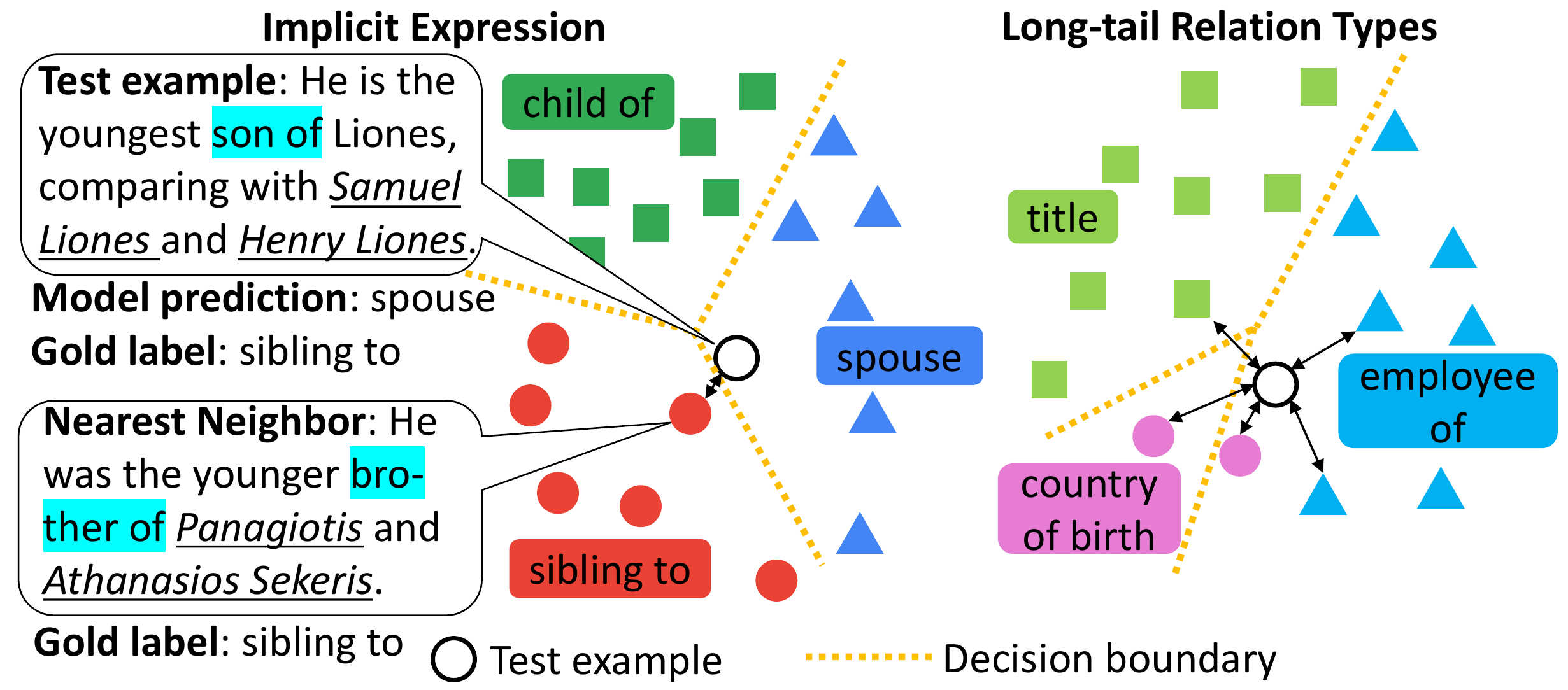}
    \caption{\textbf{Left}: the retrieved example has a similar structure but with the phrase ``younger brother'', it becomes easier to infer. 
    \textbf{Right}: Referring to the gold labels of nearest neighbors can reduce the bias.
    Highlighted words may directly influence on the relation prediction.
    }
    \label{fig:problems}
\end{figure}

\textbf{Implicit expression} refers to the situation where
a relation is expressed as the underlying message that is not explicitly stated or shown.
For example, for the relation ``sibling to'', a common expression can be ``\textit{\underline{He}} has a brother \textit{\underline{James}}'', while an implicit expression could be ``He is the youngest son of Liones,  
comparing with \textit{\underline{Samuel Liones}} and \textit{\underline{Henry Liones}}.'' In the latter case, the relation ``sibling to'' between ``\textit{\underline{Samuel Liones}}'' and ``\textit{\underline{Henry Liones}}'' is not directly expressed but could be inferred from them both are brothers of the same person. Such underlying message can easily confuse the relation classifier.  
The problem of \textbf{long-tail relation types} is caused by data sparsity in training. For example, the widely used supervised RE dataset TACRED~\cite{zhang-etal-2017-position} 
includes 41 relation types. The most frequent type ``per:title'' has 3,862 training examples, while over 22 types have less than 300 examples. The majority types
can easily dominate  model 
predictions
and lead to low performance on long-tail types.



\begin{figure*}[t]
    \centering
    \includegraphics[width=\linewidth]{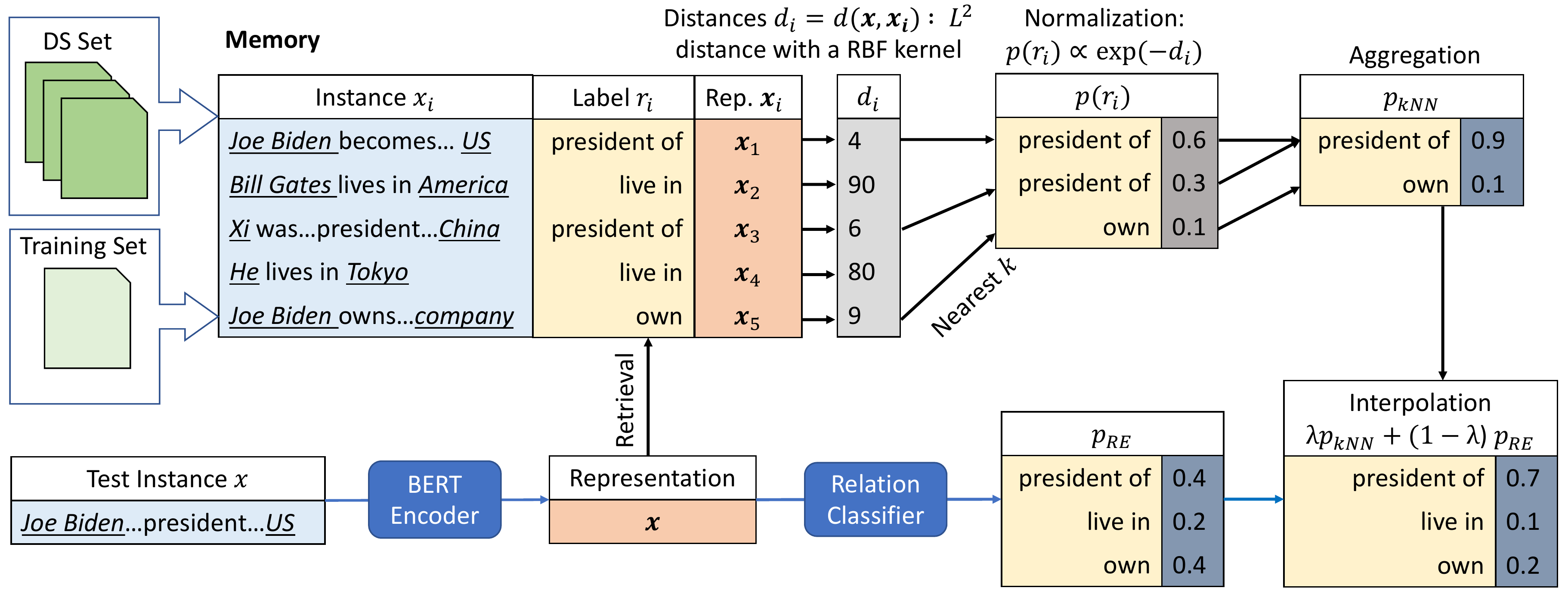}
    \caption{\textbf{An illustration of $k$NN-RE}. The memory is constructed with each pair of relation representations (Rep.) and relation labels from training set or DS set. For inference, the blue line denotes the workflow for vanilla RE and the black line denotes the workflow for $k$NN.}
    \label{fig:overview}
\end{figure*}

Inspired by recent studies~\cite{DBLP:conf/iclr/KhandelwalLJZL20, DBLP:journals/corr/abs-2002-08909,DBLP:journals/corr/abs-2110-08743} using $k$NN to retrieve diverse expressions for
language generation tasks, we introduce a simple but effective $k$NN-RE framework to address above-mentioned two problems.
Specifically, we store the training examples as the memory by a vanilla RE model and consult the stored memory at test time through a nearest-neighbor search.
As shown in Figure~\ref{fig:problems}, for an \textbf{implicit expression}, the expression ``son of'' may mislead to an incorrect prediction while its retrieved nearest neighbor contains a direct expression ``brother of'', which is a more explicit expression of the gold label ``sibling to''.
The prediction of \textbf{long-tail} examples, as shown in Figure~\ref{fig:problems}, is usually biased toward the majority class. Nearest neighbor retrieval provides direct guidance to the prediction by 
referring to the labels of its nearest neighbors in the training set, and thus can significantly reduce the imbalanced classification.


Additionally, we observe that $k$NN-RE serves as an efficient way to leverage distant supervision (DS)  data for RE. DS augments labeled RE datasets by matching knowledge base (KB) relation triplets and raw text entity pairs in a weak-supervision fashion  ~\cite{mintz-etal-2009-distant,lin-etal-2016-neural,vashishth-etal-2018-reside,chen-etal-2021-cil}.  
Recent studies~\cite{baldini-soares-etal-2019-matching,DBLP:journals/corr/abs-2102-09681,peng-etal-2020-learning,https://doi.org/10.48550/arxiv.2205.08770}, which apply PLMs to the DS labeled data 
 to improve supervised RE,  
 require heavy computation due to the fact that 
they require pre-training on DS data, whose size is usually dozens of times that of supervised datasets.  
To address this issue, 
we propose a lightweight method to leverage DS data to benefit supervised RE by extending the construction of stored memory for $k$NN-RE to DS labeled data and outperforming the recent best pre-training method with no extra training. 


In summary, we propose $k$NN-RE: a flexible $k$NN framework to solve the RE task. We conduct the experiments for $k$NN-RE with three different memory settings: training, DS, and the combination of training and DS. The results show that our $k$NN-RE with the training memory obtains a 0.84\%-1.15\% absolute F1 improvement on five datasets and achieves state-of-the-art (SOTA) F1 scores on three of them (ACE05, SciERC and Wiki80). In the DS setup, $k$NN-RE outperforms SOTA DS pre-training methods on two datasets (i2b2, Wiki20) significantly without extra training.


\section{Methodology} \label{method}
\subsection{Background: Vanilla RE model}
For the vanilla RE model, We follow the recent SOTA method PURE~\cite{zhong-chen-2021-frustratingly}. To encode an input example to a fixed-length representation by fine-tuning PLMs such as BERT~\cite{devlin-etal-2019-bert}, PURE adds extra marker tokens to highlight the head and tail entities and their types.



\begin{table}
    \centering
    \resizebox{0.9\linewidth}{!}{
    \begin{tabular}{lrrrr}
    \toprule
        Dataset & \# Rel. & \# Train & \# Dev & \# Test \\
        \hline
        ACE05 & 6 & 4,788& 1,131 & 1,151 \\
        Wiki80 & 80 & 45,330 & 5,070 & 5,600 \\
        TACRED & 41 & 68,124 & 22,631 & 15,509 \\
        i2b2 2010VA & 8 & 3,020 & 111 & 6,147 \\
        SciERC & 7 & 1,861 & 275 & 551 \\
        \hline
        Wiki20m & 80 & 303K & - & - \\
        MIMIC-III & 8 & 36K & - & -\\
        \bottomrule
    \end{tabular}
    }
    \caption{\textbf{Statistics of datasets}. Rel. denotes relation types.}
    \label{supervised data}
\end{table}

\begin{table*}[t]
    \centering
    \resizebox{\linewidth}{!}{
    \begin{tabular}{llllll}
    \toprule
        Methods& ACE05 & Wiki80 & TACRED & i2b2 2010VA& SciERC \\
         \toprule
         \multicolumn{6}{c}{\textit{Baselines}} \\
         \hline
         ~\cite{DBLP:conf/bionlp/PengYL19}& - & - & - & \textbf{76.2$\dagger$} & - \\
         ~\cite{han-etal-2019-opennre} &- & 86.61 & - & - & - \\
         ~\cite{DBLP:journals/corr/abs-2102-01373} &- & - & \textbf{71.5} & - & -\\
         CP~\cite{peng-etal-2020-learning}$\clubsuit$ &- & 87.50 & - & 72.84 & -\\
         PURE~\cite{zhong-chen-2021-frustratingly} &74.00 & 86.70 & 69.42 & 72.28 & 68.45 \\
         \toprule
         \multicolumn{6}{c}{\textit{Ours (Best $k,\lambda$)}} \\
         \hline
         $k$NN only: Train memory & \textbf{75.07 ($4,1.0$)} & 87.35 ($4,1.0$)& 70.21 ($8,1.0$)& 73.18 ($32,1.0$) & 68.58 ($64,1.0$)\\
         $k$NN-RE: Train memory & \textbf{75.07 ($4,1.0$)} & \textbf{87.54 ($4,0.5$)} & 70.57 ($8,0.4$) & 73.38 ($32,0.7$) & \textbf{69.47 ($64,0.6$)} \\
         \toprule
         
         $k$NN-RE: DS memory $\clubsuit$ & - & 87.79 ($256,0.5$)& -& 73.22 ($64,0.3$)& -\\
         $k$NN-RE: Combined memory $\clubsuit$ & - & \textbf{88.32 ($\alpha=0.5$)}& -& 73.67 ($\alpha=0.6$) & -\\
         \bottomrule
    \end{tabular}
    }
    \caption{\textbf{Main Results of $k$NN-RE with different memory settings on five datasets}. $\clubsuit$ denotes the methods using DS set.
    $\dagger$: SOTA i2b2 2010VA adopts specific encoding. \textcolor{black}{“$k$NN only” means only using $p_{kNN}(y|x)$ and is described by $\lambda = 1$ in the parameters.
    ``Combined'' means the combination of both memories by: $\alpha p_{\mathsmaller{kNN-RE}}(Train)+(1-\alpha)p_{\mathsmaller{kNN-RE}}(DS)$, where $p_{\mathsmaller{kNN-RE}}(Train)$ and $p_{\mathsmaller{kNN-RE}}(DS)$ is computed by Equation~\ref{interpolation} corresponding to the “Train memory” and “DS memory.”, and $k$, $\lambda$ are given by the best setting of each single memory. }}
    \label{main results}
\end{table*}

\begin{table}[t]
    \centering
    \resizebox{\linewidth}{!}{
    \begin{tabular}{lll}
    \toprule
        Methods&  Wiki80 & i2b2 2010VA \\
         \toprule
         \multicolumn{3}{c}{\textit{Baselines}} \\
         \hline
         CP~\cite{peng-etal-2020-learning}$\clubsuit$ &87.32 & 75.62\\
         PURE~\cite{zhong-chen-2021-frustratingly} &85.78 & 73.45 \\
         \toprule
         \multicolumn{3}{c}{\textit{Ours (Best $k,\lambda$)}} \\
         \hline
         $k$NN only: Train memory & 86.70 ($4,1.0$) & 74.70 ($32,1.0$)\\
         $k$NN-RE: Train memory & \textbf{87.12 ($4,0.5$)} & \textbf{75.20 ($32,0.7$)}  \\
         \toprule
         
         $k$NN-RE: DS memory$\clubsuit$ & 88.20($256,0.5$) & 76.80 ($64,0.3$)\\
         $k$NN-RE: Combined memory$\clubsuit$ & \textbf{88.54 ($\alpha=0.5$)} & \textbf{78.25 ($\alpha=0.6$)}\\
         \bottomrule
    \end{tabular}
    }
    \caption{\textbf{\textcolor{black}{Results on development set. }}. $\clubsuit$ denotes the methods using DS set.}
    \label{Dev results}
\end{table}

Specifically, given an example $x$: ``\textit{\underline{He}} has a brother \textit{\underline{James}}.'', the input sequence is ``\textsf{\small [CLS] [H\_PER]} \textit{\underline{He}} \textsf{\small [/H\_PER]} has a brother \textsf{\small [T\_PER]} \textit{\underline{James}} \textsf{\small [/T\_PER]}. \textsf{\small [SEP]}'' where ``PER'' is the entity type if provided.
Denote the $n$-th hidden representation of the BERT encoder as $\mathbf{h}_n$. Assuming $i$ and $j$ are the indices of two beginning entity markers \textsf{\small [H\_PER]} and \textsf{\small [T\_PER]}, we define the relation representation as $\mathbf{x}= \mathbf{h}_i \oplus \mathbf{h}_j$ where $\oplus$ stands for concatenation. Subsequently, this representation is fed into a linear layer to generate the probability distribution $p_{\mathsmaller{RE}}(y|x)$ for predicting the relation type.


\subsection{Proposed Method: $k$NN-RE}
\paragraph{Training Memory Construction}

For the $i$-th training example $(x_i,r_i)$, we construct the \textit{key-value} pair $(\mathbf{x_i}, r_i)$ where the \textit{key} $ \mathbf{x_i}$ is the relation representation obtained from the vanilla RE model and the \textit{value} $r_i$ denotes the labeled relation type. The memory $(\mathcal{K}, \mathcal{V})=\{(\mathbf{x_i}, r_i)|(x_i,r_i)\in \mathcal{D}\} $ is thus the set of all \textit{key-value} pairs constructed from all the labeled examples in the training set $\mathcal{D}$.

\paragraph{DS Memory Construction}
In this paper, with the awareness of the unique feature of RE to generate abundant labeled data by DS, we extend our method by leveraging DS examples for memory construction.
Similar to training memory construction, we build \textit{key-value} pairs for all the DS labeled examples with the vanilla RE model.

\paragraph{Inference}
Given the test example $x$, the RE model outputs its relation representation $\mathbf{x}$ and generate the relation distribution  $ p_{\mathsmaller{RE}}(y|x)$ between two mentioned entities. We then query the memory with $\mathbf{x}$ to retrieve its $k$ nearest neighbors $\mathcal{N}$ according to a distance function $d(.,.)$ by $L^{2}$ distance with the KBF kernel. We weight retrieved examples by a softmax function on the negative distance and make an aggregation on the labeled relation types to predict a relation distribution $ p_{\mathsmaller{kNN}}(y|x)$: 
\begin{equation}
p_{\mathsmaller{kNN}}(y|x) \propto \sum_{(\mathbf{x_i}, r_i)\in \mathcal{N}}\mathbbm{1}_{y=r_i} \frac{exp(-d(\mathbf{x},\mathbf{x_i}))}{\mathcal{T}}
\label{kNN_predict}
\end{equation}
where $\mathcal{T}$ denotes a scaling temperature.
Finally, we interpolate the RE model distribution $ p_{\mathsmaller{RE}}(y|x)$ and $k$NN distribution $ p_{\mathsmaller{kNN}}(y|x)$ to produce the final overall distribution:
\begin{equation}
p_{\mathsmaller{kNN-RE}}(y|x)= \lambda p_{\mathsmaller{kNN}}(y|x)+(1-\lambda)p_{\mathsmaller{RE}}(y|x)
\label{interpolation}
\end{equation}
where $\lambda$ is a hyperparameter.

\section{Experiment settings} \label{experiment}
\paragraph{Supervised Datasets}
We evaluate our proposed method on five popular RE datasets. Table~\ref{supervised data} shows the statistics. ACE05 and TACRED datasets are built over an assortment of newswire and online text. Wiki80~\cite{han-etal-2019-opennre} is derived from Wikipedia crossing various domains. The i2b2 2010VA dataset collects medical reports while SciERC~\cite{luan-etal-2018-multi} collects AI paper abstracts and annotated relations, specially for scientific knowledge graph construction.

\paragraph{DS Datasets}
We evaluate our DS memory construction method on two supervised datasets Wiki80 and i2b2 2010VA. For i2b2 2010VA, we generate DS examples from MIMIC-III based on triplets extracted from the training set. For Wiki80 dataset, as it is derived from the Wikidata KB, we leverage the existing DS dataset Wiki20m derived from the same KB which leads to shared relation types.

Refer to Appendix~\ref{plm} for implementation details.


\begin{table}[t]\Large
    \centering
    \resizebox{\linewidth}{!}{
    \begin{tabular}{l|c}
    \toprule
        \textbf{Test example $1$} & $\mathbf{p_{\mathsmaller{RE}}}$\\
        \toprule
        He was the captain of ... team that won the& \multirow{3}{*}{ \makecell[c]{part of: $0.3$ ($\surd$)\\participant of: $0.7$}} \\
        \textit{\underline{1950 World Cup}} after beating ... in the final & \\
        round match known as the "\textit{\underline{Maracanazo}}".&  \\
        \bottomrule
        \textbf{Retrieved nearest neighbors}& \makecell[c]{\textbf{Gold label}\\$\mathbf{d_i\rightarrow p(r_i)}$}\\
        \toprule
        \textit{\underline{Turkish Cypriots}} are the minority of the island &\multirow{2}{*}{\makecell[c]{part of \\ $1\rightarrow 0.9$}}\\
        with \textit{\underline{Turkish Settlers}} from Turkey and ...
        &  \\

        \bottomrule
        It was succeeded as Danish representative at  & \multirow{2}{*}{\makecell[c]{participant of\\$5 \rightarrow 0.03$}}\\
        \textit{\underline{2006 Contest}} by \textit{\underline{Sidsel Ben Semmane}} with ... & \\
        \bottomrule
        
    \toprule
        \textbf{Test example $2$} & $\mathbf{p_{\mathsmaller{RE}}}$\\
        \toprule
        \textit{\underline{South Air Command}} of the \textit{\underline{Indian Air Force}}& \multirow{2}{*}{ \makecell[c]{subsidiary: $0.478$ ($\surd$)\\has part: $0.496$}} \\
         is headquartered in the city & \\

        \bottomrule
        \textbf{Retrieved nearest neighbors}& \makecell[c]{\textbf{Gold label}\\$\mathbf{d_i\rightarrow p(r_i)}$}\\
        \toprule
        The artwork depicts the Christian \textit{\underline{Holy  Family}}&\multirow{3}{*}{\makecell[c]{has part \\ $2\rightarrow 0.6$}}\\
         of \textit{\underline{Mary}}, Joseph, and the infant Jesus, in an
        &  \\
         enclosed garden, symbolizing Mary 's virginity.
        &  \\

        \bottomrule
        Fidobank (formerly \textit{\underline{SEB Bank}}) is a bank of Ukraine   & \multirow{2}{*}{\makecell[c]{subsidiary\\$3 \rightarrow 0.3$}}\\
        that until 2012 belonged to the Swedish \textit{\underline{SEB Group}}. & \\
        \bottomrule
    \end{tabular}
    }
    \caption{\textbf{\textcolor{black}{Two implicit test examples from Wiki80}}.}
    \label{implicit expression}
\end{table}

\begin{table}[t]
    \centering
    \resizebox{\linewidth}{!}{
    \begin{tabular}{lrrrc}
    \toprule
        Relation type & \# Train & \# Test& PURE& $k$NN-RE\\
         \hline
         per:charges & 280 & 103 & 75.14 & \textbf{77.23} (+2.09) \\
         org:founded by & 268 & 68 & 64.06& \textbf{64.12} (+0.06) \\
         per:siblings & 250 & 55 & 71.84& \textbf{72.90} (+1.06) \\
         per:s\_a & 229 & 30 & 64.00 & \textbf{66.67} (+2.67) \\
         per:c\_of\_d& 227& 28 & 25.00 & \textbf{30.30} (+5.30) \\
         org:founded & 166 & 37 & 80.25 & \textbf{82.05} (+1.80) \\
         per:religion & 153 & 47 & 65.82 & \textbf{68.24} (+2.42) \\
         org:p/r\_a & 125 & 10 & 47.62 & \textbf{50.00} (+3.38) \\
         org:n\_of\_e/m & 121 & 19 & 73.68 & \textbf{74.29} (+0.61) \\      per:s\_o\_b & 61& 8 & \textbf{66.67} & 61.54 (-5.13) \\
         \bottomrule
    \end{tabular}
    }
    \caption{\textbf{Results on long-tail relation types}. s\_a: schools attended, c\_of\_d: city of death, p/r\_a: political/religious affiliation, n\_of\_e/m: number of employees/members, s\_o\_b: stateorprovince of birth,.}
    \label{long-tail types}
\end{table}



\section{Results and Analysis}
\subsection{Main results}
Table~\ref{main results} compares our proposed $k$NN-RE with previous SOTA methods. For training memory construction, we can observe that: (1) Our approach outperforms vanilla RE model PURE on all datasets and achieves new SOTA performances on ACE05, Wiki80 and SciERC (2) By comparing PURE and ``$k$NN only'', we find that the $k$NN prediction itself has a performance improvement over the vanilla RE model from $+0.13$ on SciERC to $+1.07$ on ACE05. This leads to a large $\lambda$ for $k$NN.
For DS and combined memory construction, we can observe that: (1) The best $\lambda$ becomes smaller while $k$ becomes larger which is reasonable as DS suffers from the noise but still benefit the model inference. (2) Compared with the previous best pre-training method CP~\cite{peng-etal-2020-learning} that requires huge computation, our $k$NN-RE on DS achieves higher performance without extra training and is further improved by combining with the training memory to achieve a $1.62$ F1 score improvement over PURE on the Wiki80.

\textcolor{black}{Besides, we also compare performances on the development set of two datasets as shown in Table~\ref{Dev results}, the experiment results emphasize the consistent improvement of our proposed methods.}
\subsection{Analysis}
\paragraph{Case Study for Implicit Expressions}
We select two typical test examples to better illustrate the amendment by $k$NN retrieval \textcolor{black}{ as shown in Table~\ref{implicit expression}}. For the first example, 
the implicit relation between ``\textit{\underline{maracanazo}}'' and ``\textit{\underline{1950 world cup}}'' need to be inferred by other contextual information and the RE model makes an incorrect prediction as competition usually belongs to the object of the relation ``participant of'' as the second retrieved examples. However, the nearest example contains a simpler expression and rectifies the prediction. Refer to Appendix~\ref{tsne} for visualized analysis.

\textcolor{black}{For the second example, the implicit expression leads to another confusing relation "subsidiary" while the nearest example captures the same structure that contains an "of" between two entities and makes the final prediction easier.}
\paragraph{Performance on Long-tail Relation Types}
We check the performance $k$NN-RE with training memory on several most long-tail relation types from the TACRED dataset and show in table~\ref{long-tail types}. Note that all long-tail relation types benefit from the effectiveness of $k$NN prediction except for ``stateorprovince of birth'', which contains only 8 test examples leading to an unconvincing performance.
\begin{figure}[t]
    \centering
    \includegraphics[width=\linewidth]{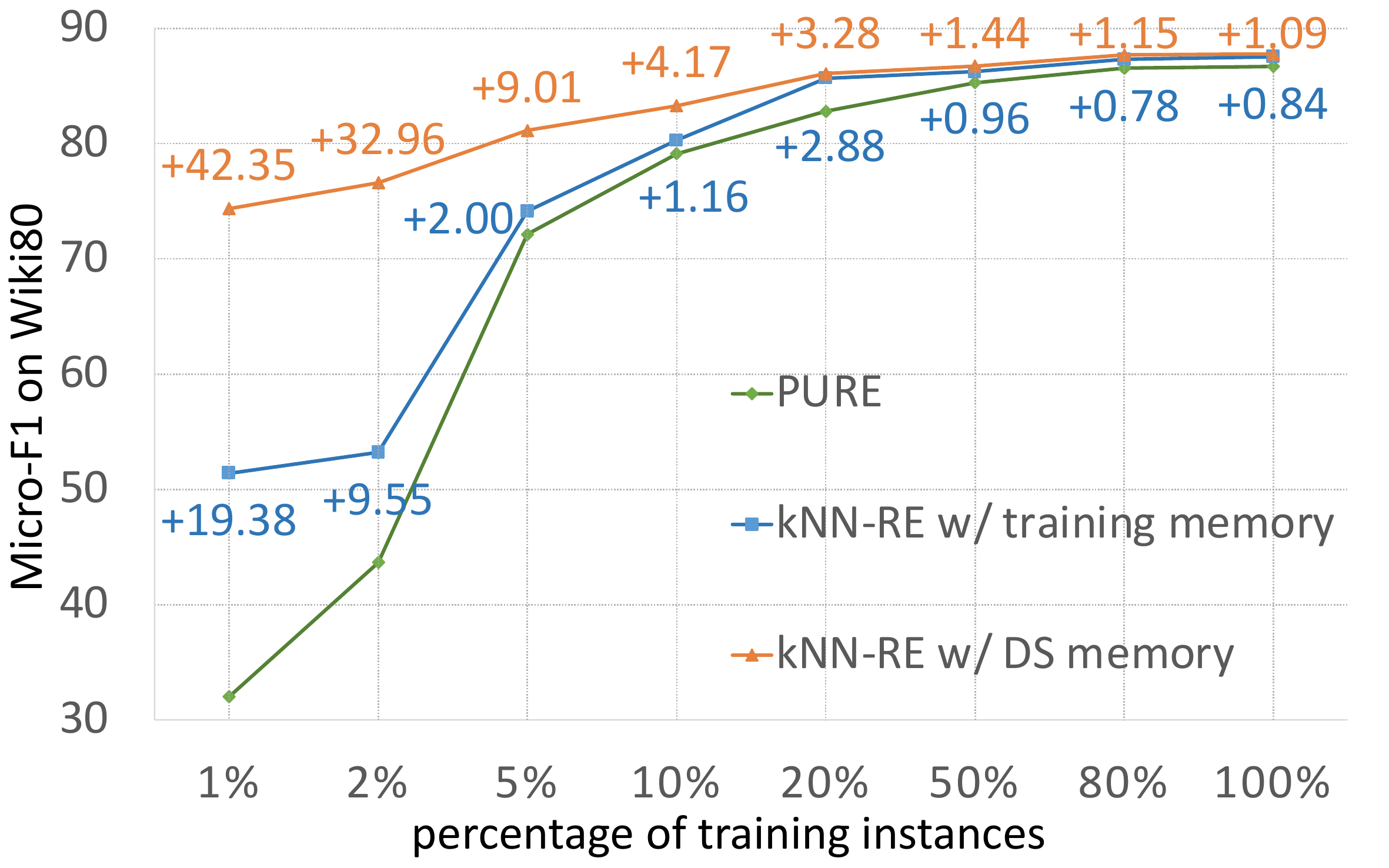}
    \caption{\textbf{Analyzing retrieval ability on Wiki80}.}
    \label{low_resource}
\end{figure}

\paragraph{Retrieval Ability in Low-Resource Scenario}
We also check the retrieval ability by varying the percentage of the training set to constraint the representation quality in the memory (Figure~\ref{low_resource}). We can observe that with the decreasing number of the training examples, our $k$NN-RE (training) tends to achieve greater improvement even the training memory is also limited by the low resource. Surprisingly, our $k$NN-RE (DS) achieves the F1-score of $74.31$ (an improvement gap of $42.35$ over PURE) with only $1\%$ training examples provided, which indicates that the model can still retrieve accurate nearest neighbors from the DS memory.
We believe this is due to the modern PLMs have learned robust representations during pre-training.
\section{Conclusion}
We propose $k$NN-RE: a flexible $k$NN framework with different memory settings for solving implicit expression and long-tail relation issues in RE. The results show that our $k$NN-RE with training memory outperforms vanilla RE model and achieves SOTA F1 scores on three datasets. In the DS setup, $k$NN-RE also outperforms SOTA DS pre-training methods significantly without extra training.

\section*{Limitations}

In this paper, we use $k$NN-based strategy in the inference stage to address the language complexity and data sparsity problem. 
It is more challenging for a model to learn the characteristics of these examples. 
While our approach is light-weighted and flexible, it cannot directly help the model to improve the classification of the implicit expression examples or long-tail relation examples types during the training stage. The representations of these examples remain coarse-grained. Incorporating the $k$NN manner strategies in the training stage by providing additional nearest neighbor references for the model could help the model learn better representations of the examples, which we leave as future work.

  \section*{Acknowledgements}
  This work is partially supported by JST SPRING Grant No.JPMJSP2110, MHLW PRISM Grant Number 21AC5001, KAKENHI Number 21H00308 and KAKENHI Number 22J13719, Japan.


\bibliography{anthology,custom}

\begin{thebibliography}{24}
\expandafter\ifx\csname natexlab\endcsname\relax\def\natexlab#1{#1}\fi

\bibitem[{Baldini~Soares et~al.(2019)Baldini~Soares, FitzGerald, Ling, and
  Kwiatkowski}]{baldini-soares-etal-2019-matching}
Livio Baldini~Soares, Nicholas FitzGerald, Jeffrey Ling, and Tom Kwiatkowski.
  2019.
\newblock \href {https://doi.org/10.18653/v1/P19-1279} {Matching the blanks:
  Distributional similarity for relation learning}.
\newblock In \emph{Proceedings of the 57th Annual Meeting of the Association
  for Computational Linguistics}, pages 2895--2905, Florence, Italy.
  Association for Computational Linguistics.

\bibitem[{Beltagy et~al.(2019)Beltagy, Lo, and
  Cohan}]{beltagy-etal-2019-scibert}
Iz~Beltagy, Kyle Lo, and Arman Cohan. 2019.
\newblock \href {https://doi.org/10.18653/v1/D19-1371} {{S}ci{BERT}: A
  pretrained language model for scientific text}.
\newblock In \emph{Proceedings of the 2019 Conference on Empirical Methods in
  Natural Language Processing and the 9th International Joint Conference on
  Natural Language Processing (EMNLP-IJCNLP)}, pages 3615--3620, Hong Kong,
  China. Association for Computational Linguistics.

\bibitem[{Chen et~al.(2021)Chen, Shi, Tang, Chen, Wu, and
  Zhuang}]{chen-etal-2021-cil}
Tao Chen, Haizhou Shi, Siliang Tang, Zhigang Chen, Fei Wu, and Yueting Zhuang.
  2021.
\newblock \href {https://doi.org/10.18653/v1/2021.acl-long.483} {{CIL}:
  Contrastive instance learning framework for distantly supervised relation
  extraction}.
\newblock In \emph{Proceedings of the 59th Annual Meeting of the Association
  for Computational Linguistics and the 11th International Joint Conference on
  Natural Language Processing (Volume 1: Long Papers)}, pages 6191--6200,
  Online. Association for Computational Linguistics.

\bibitem[{Cheng et~al.(2020)Cheng, Asahara, Kobayashi, and
  Kurohashi}]{cheng-etal-2020-dynamically}
Fei Cheng, Masayuki Asahara, Ichiro Kobayashi, and Sadao Kurohashi. 2020.
\newblock \href {https://doi.org/10.18653/v1/2020.findings-emnlp.121}
  {Dynamically updating event representations for temporal relation
  classification with multi-category learning}.
\newblock In \emph{Findings of the Association for Computational Linguistics:
  EMNLP 2020}, pages 1352--1357, Online. Association for Computational
  Linguistics.

\bibitem[{Devlin et~al.(2019)Devlin, Chang, Lee, and
  Toutanova}]{devlin-etal-2019-bert}
Jacob Devlin, Ming-Wei Chang, Kenton Lee, and Kristina Toutanova. 2019.
\newblock \href {https://doi.org/10.18653/v1/N19-1423} {{BERT}: Pre-training of
  deep bidirectional transformers for language understanding}.
\newblock In \emph{Proceedings of the 2019 Conference of the North {A}merican
  Chapter of the Association for Computational Linguistics: Human Language
  Technologies, Volume 1 (Long and Short Papers)}, pages 4171--4186,
  Minneapolis, Minnesota. Association for Computational Linguistics.

\bibitem[{Guu et~al.(2020)Guu, Lee, Tung, Pasupat, and
  Chang}]{DBLP:journals/corr/abs-2002-08909}
Kelvin Guu, Kenton Lee, Zora Tung, Panupong Pasupat, and Ming{-}Wei Chang.
  2020.
\newblock \href {http://arxiv.org/abs/2002.08909} {{REALM:} retrieval-augmented
  language model pre-training}.
\newblock \emph{CoRR}, abs/2002.08909.

\bibitem[{Han et~al.(2019)Han, Gao, Yao, Ye, Liu, and
  Sun}]{han-etal-2019-opennre}
Xu~Han, Tianyu Gao, Yuan Yao, Deming Ye, Zhiyuan Liu, and Maosong Sun. 2019.
\newblock \href {https://doi.org/10.18653/v1/D19-3029} {{O}pen{NRE}: An open
  and extensible toolkit for neural relation extraction}.
\newblock In \emph{Proceedings of the 2019 Conference on Empirical Methods in
  Natural Language Processing and the 9th International Joint Conference on
  Natural Language Processing (EMNLP-IJCNLP): System Demonstrations}, pages
  169--174, Hong Kong, China. Association for Computational Linguistics.

\bibitem[{Khandelwal et~al.(2020)Khandelwal, Levy, Jurafsky, Zettlemoyer, and
  Lewis}]{DBLP:conf/iclr/KhandelwalLJZL20}
Urvashi Khandelwal, Omer Levy, Dan Jurafsky, Luke Zettlemoyer, and Mike Lewis.
  2020.
\newblock \href {https://openreview.net/forum?id=HklBjCEKvH} {Generalization
  through memorization: Nearest neighbor language models}.
\newblock In \emph{8th International Conference on Learning Representations,
  {ICLR} 2020, Addis Ababa, Ethiopia, April 26-30, 2020}. OpenReview.net.

\bibitem[{Lin et~al.(2016)Lin, Shen, Liu, Luan, and Sun}]{lin-etal-2016-neural}
Yankai Lin, Shiqi Shen, Zhiyuan Liu, Huanbo Luan, and Maosong Sun. 2016.
\newblock \href {https://doi.org/10.18653/v1/P16-1200} {Neural relation
  extraction with selective attention over instances}.
\newblock In \emph{Proceedings of the 54th Annual Meeting of the Association
  for Computational Linguistics (Volume 1: Long Papers)}, pages 2124--2133,
  Berlin, Germany. Association for Computational Linguistics.

\bibitem[{Lin et~al.(2020)Lin, Ji, Huang, and Wu}]{lin-etal-2020-joint}
Ying Lin, Heng Ji, Fei Huang, and Lingfei Wu. 2020.
\newblock \href {https://doi.org/10.18653/v1/2020.acl-main.713} {A joint neural
  model for information extraction with global features}.
\newblock In \emph{Proceedings of the 58th Annual Meeting of the Association
  for Computational Linguistics}, pages 7999--8009, Online. Association for
  Computational Linguistics.

\bibitem[{Luan et~al.(2018)Luan, He, Ostendorf, and
  Hajishirzi}]{luan-etal-2018-multi}
Yi~Luan, Luheng He, Mari Ostendorf, and Hannaneh Hajishirzi. 2018.
\newblock \href {https://doi.org/10.18653/v1/D18-1360} {Multi-task
  identification of entities, relations, and coreference for scientific
  knowledge graph construction}.
\newblock In \emph{Proceedings of the 2018 Conference on Empirical Methods in
  Natural Language Processing}, pages 3219--3232, Brussels, Belgium.
  Association for Computational Linguistics.

\bibitem[{Meng et~al.(2021)Meng, Zong, Li, Sun, Zhang, Wu, and
  Li}]{DBLP:journals/corr/abs-2110-08743}
Yuxian Meng, Shi Zong, Xiaoya Li, Xiaofei Sun, Tianwei Zhang, Fei Wu, and Jiwei
  Li. 2021.
\newblock \href {http://arxiv.org/abs/2110.08743} {{GNN-LM:} language modeling
  based on global contexts via {GNN}}.
\newblock \emph{CoRR}, abs/2110.08743.

\bibitem[{Mintz et~al.(2009)Mintz, Bills, Snow, and
  Jurafsky}]{mintz-etal-2009-distant}
Mike Mintz, Steven Bills, Rion Snow, and Daniel Jurafsky. 2009.
\newblock \href {https://aclanthology.org/P09-1113} {Distant supervision for
  relation extraction without labeled data}.
\newblock In \emph{Proceedings of the Joint Conference of the 47th Annual
  Meeting of the {ACL} and the 4th International Joint Conference on Natural
  Language Processing of the {AFNLP}}, pages 1003--1011, Suntec, Singapore.
  Association for Computational Linguistics.

\bibitem[{Orm{\'{a}}ndi et~al.(2021)Orm{\'{a}}ndi, Saleh, Winter, and
  Rao}]{DBLP:journals/corr/abs-2102-09681}
R{\'{o}}bert Orm{\'{a}}ndi, Mohammad Saleh, Erin Winter, and Vinay Rao. 2021.
\newblock \href {http://arxiv.org/abs/2102.09681} {Webred: Effective
  pretraining and finetuning for relation extraction on the web}.
\newblock \emph{CoRR}, abs/2102.09681.

\bibitem[{Peng et~al.(2020)Peng, Gao, Han, Lin, Li, Liu, Sun, and
  Zhou}]{peng-etal-2020-learning}
Hao Peng, Tianyu Gao, Xu~Han, Yankai Lin, Peng Li, Zhiyuan Liu, Maosong Sun,
  and Jie Zhou. 2020.
\newblock \href {https://doi.org/10.18653/v1/2020.emnlp-main.298} {{L}earning
  from {C}ontext or {N}ames? {A}n {E}mpirical {S}tudy on {N}eural {R}elation
  {E}xtraction}.
\newblock In \emph{Proceedings of the 2020 Conference on Empirical Methods in
  Natural Language Processing (EMNLP)}, pages 3661--3672, Online. Association
  for Computational Linguistics.

\bibitem[{Peng et~al.(2019)Peng, Yan, and Lu}]{DBLP:conf/bionlp/PengYL19}
Yifan Peng, Shankai Yan, and Zhiyong Lu. 2019.
\newblock \href {https://doi.org/10.18653/v1/w19-5006} {Transfer learning in
  biomedical natural language processing: An evaluation of {BERT} and elmo on
  ten benchmarking datasets}.
\newblock In \emph{Proceedings of the 18th BioNLP Workshop and Shared Task,
  BioNLP@ACL 2019, Florence, Italy, August 1, 2019}, pages 58--65. Association
  for Computational Linguistics.

\bibitem[{Vashishth et~al.(2018)Vashishth, Joshi, Prayaga, Bhattacharyya, and
  Talukdar}]{vashishth-etal-2018-reside}
Shikhar Vashishth, Rishabh Joshi, Sai~Suman Prayaga, Chiranjib Bhattacharyya,
  and Partha Talukdar. 2018.
\newblock \href {https://doi.org/10.18653/v1/D18-1157} {{RESIDE}: Improving
  distantly-supervised neural relation extraction using side information}.
\newblock In \emph{Proceedings of the 2018 Conference on Empirical Methods in
  Natural Language Processing}, pages 1257--1266, Brussels, Belgium.
  Association for Computational Linguistics.

\bibitem[{Wan et~al.(2022)Wan, Cheng, Liu, Mao, Song, and
  Kurohashi}]{https://doi.org/10.48550/arxiv.2205.08770}
Zhen Wan, Fei Cheng, Qianying Liu, Zhuoyuan Mao, Haiyue Song, and Sadao
  Kurohashi. 2022.
\newblock \href {https://doi.org/10.48550/ARXIV.2205.08770} {Relation
  extraction with weighted contrastive pre-training on distant supervision}.

\bibitem[{Wang and Lu(2020)}]{wang-lu-2020-two}
Jue Wang and Wei Lu. 2020.
\newblock \href {https://doi.org/10.18653/v1/2020.emnlp-main.133} {Two are
  better than one: Joint entity and relation extraction with table-sequence
  encoders}.
\newblock In \emph{Proceedings of the 2020 Conference on Empirical Methods in
  Natural Language Processing (EMNLP)}, pages 1706--1721, Online. Association
  for Computational Linguistics.

\bibitem[{Zeng et~al.(2020)Zeng, Zhang, and Liu}]{Zeng_Zhang_Liu_2020}
Daojian Zeng, Haoran Zhang, and Qianying Liu. 2020.
\newblock \href {https://doi.org/10.1609/aaai.v34i05.6495} {Copymtl: Copy
  mechanism for joint extraction of entities and relations with multi-task
  learning}.
\newblock \emph{Proceedings of the AAAI Conference on Artificial Intelligence},
  34(05):9507--9514.

\bibitem[{Zhang et~al.(2020)Zhang, Liu, Fan, Ji, Zeng, Cheng, Kawahara, and
  Kurohashi}]{zhang-etal-2020-minimize}
Ranran~Haoran Zhang, Qianying Liu, Aysa~Xuemo Fan, Heng Ji, Daojian Zeng, Fei
  Cheng, Daisuke Kawahara, and Sadao Kurohashi. 2020.
\newblock \href {https://doi.org/10.18653/v1/2020.findings-emnlp.23} {Minimize
  exposure bias of {S}eq2{S}eq models in joint entity and relation extraction}.
\newblock In \emph{Findings of the Association for Computational Linguistics:
  EMNLP 2020}, pages 236--246, Online. Association for Computational
  Linguistics.

\bibitem[{Zhang et~al.(2017)Zhang, Zhong, Chen, Angeli, and
  Manning}]{zhang-etal-2017-position}
Yuhao Zhang, Victor Zhong, Danqi Chen, Gabor Angeli, and Christopher~D.
  Manning. 2017.
\newblock \href {https://doi.org/10.18653/v1/D17-1004} {Position-aware
  attention and supervised data improve slot filling}.
\newblock In \emph{Proceedings of the 2017 Conference on Empirical Methods in
  Natural Language Processing}, pages 35--45, Copenhagen, Denmark. Association
  for Computational Linguistics.

\bibitem[{Zhong and Chen(2021)}]{zhong-chen-2021-frustratingly}
Zexuan Zhong and Danqi Chen. 2021.
\newblock \href {https://doi.org/10.18653/v1/2021.naacl-main.5} {A
  frustratingly easy approach for entity and relation extraction}.
\newblock In \emph{Proceedings of the 2021 Conference of the North American
  Chapter of the Association for Computational Linguistics: Human Language
  Technologies}, pages 50--61, Online. Association for Computational
  Linguistics.

\bibitem[{Zhou and Chen(2021)}]{DBLP:journals/corr/abs-2102-01373}
Wenxuan Zhou and Muhao Chen. 2021.
\newblock \href {http://arxiv.org/abs/2102.01373} {An improved baseline for
  sentence-level relation extraction}.
\newblock \emph{CoRR}, abs/2102.01373.

\end{thebibliography}
\bibliographystyle{acl_natbib}

\clearpage

\appendix
\section{T-SNE Visualization}\label{tsne}
\begin{figure}[t]
    \centering
    \includegraphics[width=\linewidth]{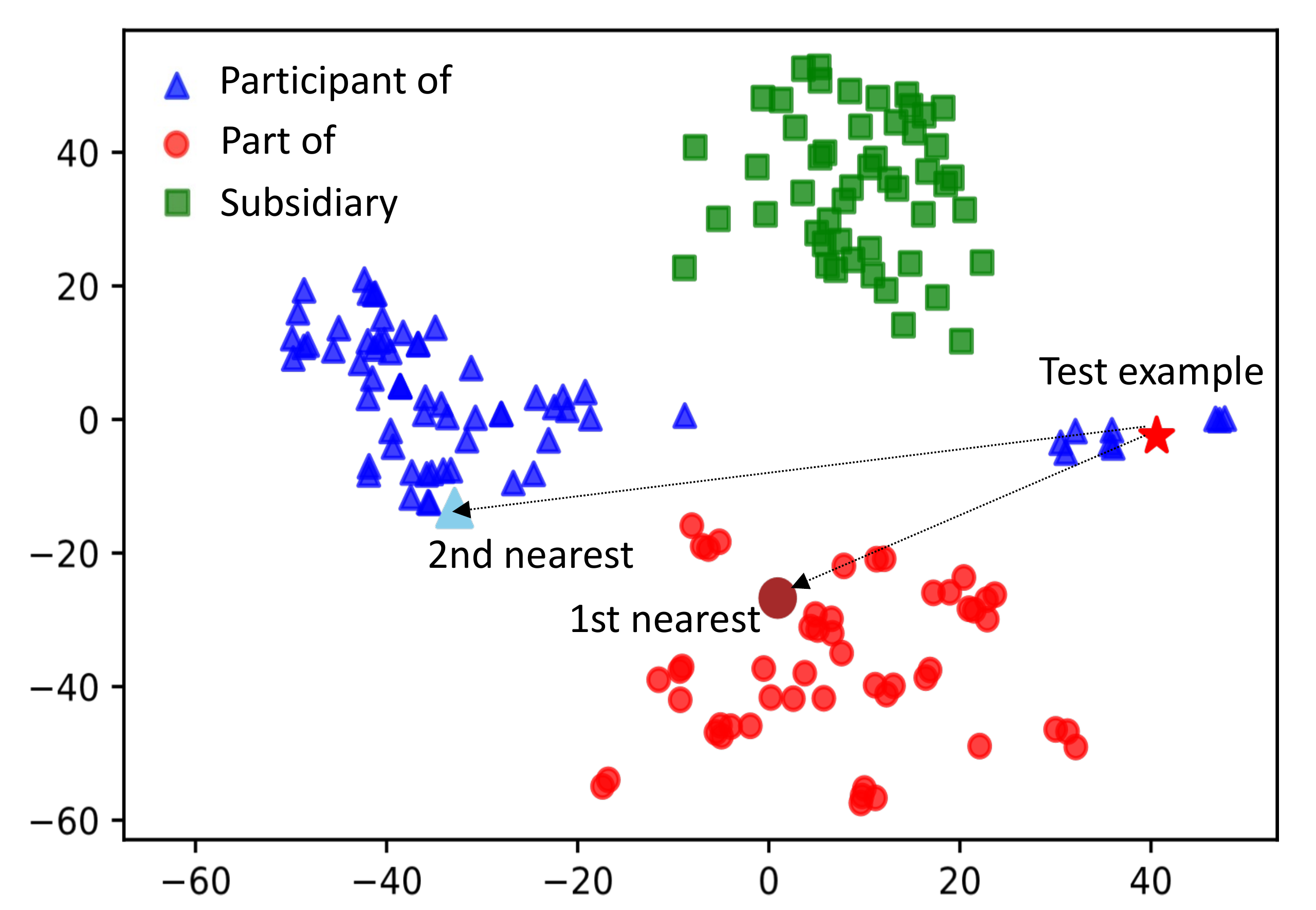}
    \caption{\textbf{The implicit test example in t-SNE}. Note that t-SNE only visualizes the distribution not the exact distance.} 
    \label{fig:tsne}
\end{figure}

As shown in figure~\ref{fig:tsne}, the test example with the gold label ``part of'' is incorrectly classified to another relation ``participant of''. However, with the help of the 1st nearest example closer to the ``part of'' clustering, $k$NN makes a correct prediction.

\section{Implementation Details}\label{plm}
During the construction of DS data for i2b2 2010VA, we use the preprocessing tool NLTK to split raw corpora into sentences.
We use \textit{bert-base-uncased}~\cite{devlin-etal-2019-bert} as the base encoders for ACE05, Wiki80 and TACRED for a fair comparison with previous work. We also use \textit{scibert-scivocab-uncased}~\cite{beltagy-etal-2019-scibert} as the base encoder for SciERC and \textit{BLUEBERT}~\cite{DBLP:conf/bionlp/PengYL19} for i2b2 2010VA as in-domain PLMs are more effective.

For the baseline PURE~\cite{zhong-chen-2021-frustratingly}, we follow their \textcolor{black}{single-sentence to keep consistency among datasets as Wiki80 and TACRED are both sentence-level RE datasets}. For another baseline CP~\cite{peng-etal-2020-learning}, we modify their official implementations to pre-train on our DS set. For our $k$NN-RE, we choose hyperparameters $k$ , $\mathcal{T}$ and $\lambda$ by greedy search where $k$ is the power of $2$ from $2$ to $256$, $\lambda$ and $\mathcal{T}$ both increase from $0$ to $1$. All experiment results are the average of 3 to 5 times running.

We used 2 NVIDIA RTX3090 for training.

\end{document}